 \documentclass[letterpaper, 10 pt, journal, twoside]{ieeetran} 

\IEEEoverridecommandlockouts                              

\usepackage{color}
\usepackage{graphics} 
\usepackage{epsfig} 
\usepackage{mathptmx} 
\usepackage{amsmath} 
\usepackage{subfigure}
\usepackage[english]{babel}
\usepackage{graphicx}
\usepackage{makecell}
\usepackage{sidecap}
\usepackage{amsmath}
\usepackage{multirow}
\usepackage{amssymb}
\usepackage{hyperref}
\usepackage{algorithm}
\usepackage{algorithmicx}
\usepackage{algpseudocode}
\title{PRIMAL$_2$: Pathfinding via Reinforcement and Imitation Multi-Agent Learning - Lifelong}

\author{Mehul Damani$^{1,*}$, Zhiyao Luo$^{1,*}$, Emerson Wenzel$^1$, Guillaume Sartoretti$^{1,\dagger}$
\thanks{Manuscript received: October, 15, 2020; Revised January, 26, 2021; Accepted February, 19, 2021.}
\thanks{This paper was recommended for publication by Editor M. Ani Hsieh
upon evaluation of the Associate Editor and Reviewers' comments.}
\thanks{$*$ These authors contributed equally to this work.}
\thanks{${\dagger}$ Corresponding author, to whom correspondence should be addressed.}
\thanks{$^{1}$ Mehul Damani, Zhiyao Luo, Emerson Wenzel, and Guillaume Sartoretti are with the department of Mechanical Engineering at the National University of Singapore, 117575 Singapore. {\tt\small damanimehul24@gmail.com, e0452733@u.nus.edu, emersonwenzel@gmail.com, mpegas@nus.edu.sg}}%
\thanks{Digital Object Identifier (DOI): see top of this page.}
} 

\begin{document}

\maketitle
\begin{abstract}

Multi-agent path finding (MAPF) is an indispensable component of large-scale robot deployments in numerous domains ranging from airport management to warehouse automation. In particular, this work addresses lifelong MAPF (LMAPF) -- an online variant of the problem where agents are immediately assigned a new goal upon reaching their current one -- in dense and highly structured environments, typical of real-world warehouse operations. Effectively solving LMAPF in such environments requires expensive coordination between agents as well as frequent replanning abilities, a daunting task for existing coupled and decoupled approaches alike. With the purpose of achieving considerable agent coordination without any compromise on reactivity and scalability, we introduce PRIMAL$_2$, a distributed reinforcement learning framework for LMAPF where agents learn fully decentralized policies to reactively plan paths online in a partially observable world. We extend our previous work, which was effective in low-density sparsely occupied worlds, to highly structured and constrained worlds by identifying behaviors and conventions which improve implicit agent coordination, and enable their learning through the construction of a novel local agent observation and various training aids. 
We present extensive results of PRIMAL$_2$ in both MAPF and LMAPF environments and compare its performance to state-of-the-art planners in terms of makespan and throughput.
We show that PRIMAL$_2$ significantly surpasses our previous work and performs comparably to these baselines, while allowing real-time re-planning and scaling up to 2048~agents.

\end{abstract}



\section{Introduction}
\label{RAL2021-section:introduction}

Multi-agent pathfinding (MAPF) is a challenging NP-hard problem with numerous real-life applications such as surveillance, search and rescue, and warehouses~\cite{nagorny2012service,berger2015innovative}.
In particular, the goal of \textit{one-shot} MAPF is to find collision-free paths for a team of agents from their start positions to their goal positions with the aim of minimizing a defined objective function, such as the \textit{makespan} (i.e., the time until all robots are on target) or the sum of their path lengths.
However, many real-world problems are dynamic and often require agents to tackle a series of targets instead of staying stationary after reaching the first one.
\textit{Lifelong} multi-agent pathfinding (LMAPF), is a variant of MAPF where agents are repeatedly assigned new goal locations and are required to reactively compute paths online~\cite{ma2017lifelong,vcap2015complete,liu2019task,nguyen2019generalized,vsvancara2019online,li2020lifelong}.
The performance of LMAPF is generally measured in terms of \textit{throughput}, i.e., the average number of targets reached per unit time.
In contrast to conventional one-shot MAPF, LMAPF, typical of factory-like environments, poses additional challenges as it requires online algorithms capable of frequently replanning as goals change.
LMAPF becomes even more challenging in densely populated, structured worlds typical of factory-like environments due to the high number of conflicts between individual agent paths.
The main contribution of this paper is the introduction of PRIMAL$_2$, a distributed reinforcement learning framework that extends our previous work in one-shot MAPF, PRIMAL~\cite{sartoretti2019primal}, to LMAPF for dense, structured warehouse-like environments.
In this new framework, special emphasis is laid on achieving extensive implicit agent coordination during lifelong MAPF for arbitrarily large team sizes, while remaining fully decentralized and relying on local interactions only.

\begin{figure}[t]
  \centering
  \includegraphics[width=0.9\linewidth]{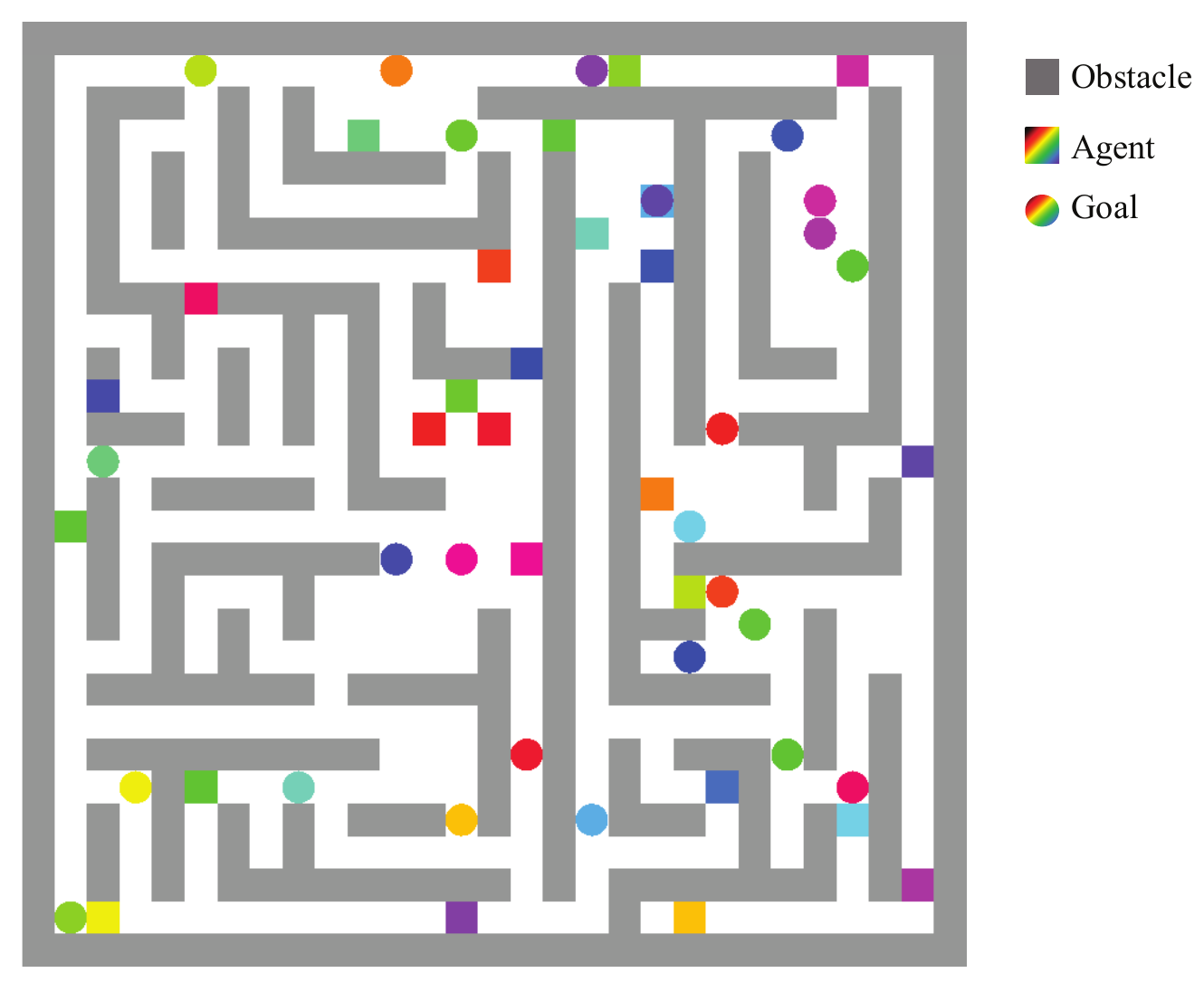}
  \vspace{-0.5cm}
  \caption{Example of the type of highly structured environments we consider.}
  \label{RAL2021-fig:map_demo}
  \vspace{-0.15cm}
\end{figure}

We focus on planning decentralized paths for a large population of agents in highly structured grid worlds, where obstacles compose narrow corridors that only allow one agent to pass at a time.
To this end, we rely on a threefold approach: first, we identify ideal behaviors and conventions that bring harmony to the movements of completely decentralized agents and enable the learning of such behavior through training aids and an observation consisting of rich feature maps.
Second, we provide agents with an intuition about future states of their surroundings, by giving each of them accesses to their neighbors' predicted future movements (using single-agent A* and ignoring other agents).
Third, and drawing from the lessons of PRIMAL, we rely on imitation learning through a centralized planner to instill favorable behaviors that are difficult to learn through vanilla RL.
We also present a new distributed training code which relies on Ray~\cite{moritz2018ray} and shows significant speed gains over our previous works, by allowing us to train models in about $12$ hours, compared to $10$ days previously.

We present results of an extensive set of simulations containing up to $2048$ agents for both one-shot MAPF and LMAPF in dense, highly-structured environments.
There, we experimentally demonstrate that PRIMAL$_2$ agents successfully learn to adhere to necessary conventions and execute coordinated manoeuvres which maximize joint performance without any explicit communication.
Our results also show that PRIMAL$_2$ is able to significantly surpass the performance of our previous work and perform on par with state-of-the-art planners in multiple scenarios, which resemble real-life multi-robot deployments in structured environments.

The paper is structured as follows: Section~\ref{RAL2021-section:prior} discusses the state-of-the-art in one-shot and lifelong MAPF.
Section~\ref{RAL2021-section:problem} formulates the specific one-shot and LMAPF problems considered.
Section~\ref{RAL2021-section:policy} proposes PRIMAL$_2$ and details the RL framework, while Section~\ref{RAL2021-section:learning} describes how learning is carried out.
Finally, Section~\ref{RAL2021-section:results} presents and discusses the results from our simulations, and Section~\ref{RAL2021-section:conclusion} contains the closing remarks.

\section{Prior Works}
\label{RAL2021-section:prior}


\subsection{One-shot Multi-Agent Pathfinding}

MAPF planners can be broadly divided into three categories: coupled, decoupled, and dynamically coupled.
Coupled planners use the high-dimensional joint space to find complete and (bounded sub)optimal paths but at a high computational cost, which scales exponentially with the number of agents~\cite{hart1968formal,standley2010finding}.
On the other hand, decoupled planners plan in the low dimensional space of each agent, and adjust paths to avoid collisions~\cite{saha2006multi,silver2005cooperative,erdmann1987multiple}.
In particular, many recent works have started looking to machine learning methods to learn decentralized policies for MAPF~\cite{sartoretti2019primal,zhang2020learning,li2020graph}.
Although significantly faster than coupled approaches, decoupled planners do not guarantee optimal solutions and are typically not complete.
Dynamically coupled approaches lie between coupled and decoupled approaches, by seeking to only increase the search space when needed~\cite{wagner2015subdimensional,sharon2015conflict}.
They are able to find (bounded sub)optimal solutions without exploring the full joint configuration space.

In particular, our recent work, PRIMAL~\cite{sartoretti2019primal}, proposed to address the trade-off between high-quality paths and scalability by relying on distributed reinforcement learning (RL) to teach agents fully decentralized reactive policies capable of computing individual paths online.
Although PRIMAL scales well to arbitrarily large team sizes, it performs poorly in structured, densely occupied worlds that require substantial agent coordination to be solved effectively.
To address this limitation of communication-free, decentralized MAPF planners, recent works have also proposed allowing agents to learn local communication and decision making policies in constrained environments using graph neural networks~\cite{li2020graph}.
However, these communication learning methods often suffer from poor scalability to larger teams.


\subsection{Lifelong Multi-Agent Pathfinding}

One of the common approaches to solve LMAPF involves stitching one-shot MAPF instances together by using a (usually complete, bounded suboptimal) MAPF planner to recompute paths at each timestep at least one agent is assigned a new goal~\cite{liu2019task,wan2018lifelong,vcap2015complete}.
However, replanning time grows exponentially with the number of agents, and resources are wasted in the redundant computation of paths for agents whose goals are unaffected.
Svancara et al.~\cite{vsvancara2019online} adapted one-shot MAPF solvers for LMPAF, which reuse paths from previous planning iterations.
However, dense, high-traffic worlds may contain many agents with conflicting paths, where significant replanning is still required.
Some planners plan new paths for only the agents that have a new goal location, but have to resort to non-optimal techniques such as holding an agent's position or providing a dummy path~\cite{ma2017lifelong}.
Another very recent and promising approach is to plan paths within a finite window, which leads to better scalability and a more reactive algorithm~\cite{li2020lifelong}, but at the cost of \textit{completeness}.
This phenomenon worsens when the planning window size is small in comparison to the average distance to goal, as agents cannot anticipate the situation outside the planning window and might plan greedy short-term paths that lead to unsolvable scenarios.

\section{Problem Formulation}
\label{RAL2021-section:problem}

\begin{figure}[b]
    \centering
    \vspace{-0.3cm}
    \subfigure[]{
        \begin{minipage}[t]{0.2\linewidth}
        \centering
        \includegraphics[height=1.75cm]{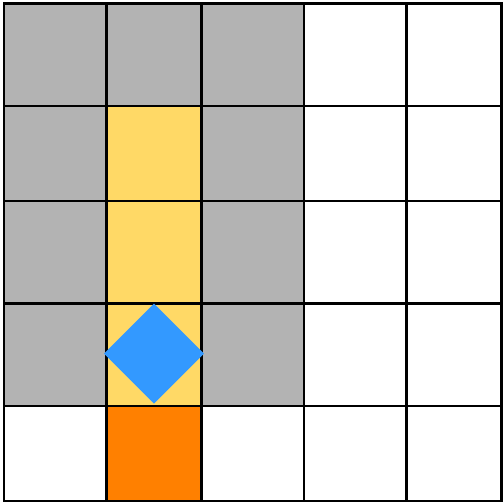}
        \end{minipage}%
    }
    \subfigure[]{
        \begin{minipage}[t]{0.2\linewidth}
        \centering
        \includegraphics[height=1.75cm]{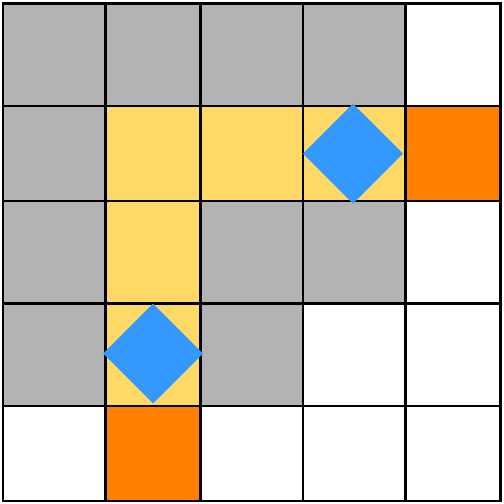}
        \end{minipage}%
    }
    \subfigure[]{
        \begin{minipage}[t]{0.5\linewidth}
        \centering
        \includegraphics[height=1.75cm]{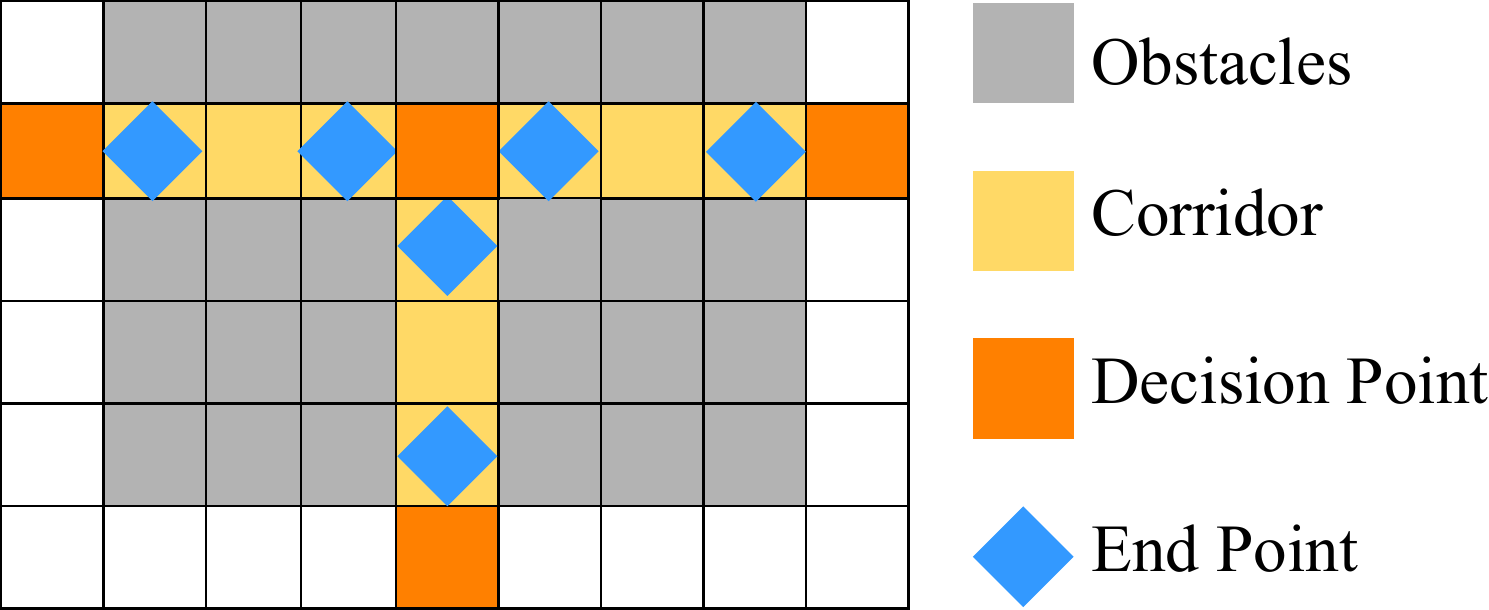}
        \end{minipage}%
    }
    \centering
    \vspace{-0.3cm}
    \caption{Corridor examples: (a) dead-end, (b) usual corridor with two endpoints, and (c) combination of 3 corridors forming a T-junction.}
    \label{RAL2021-fig:corridor}
    \vspace{-0.02cm}
\end{figure}

\begin{figure*}
  \centering
  \includegraphics[width=0.96\linewidth]{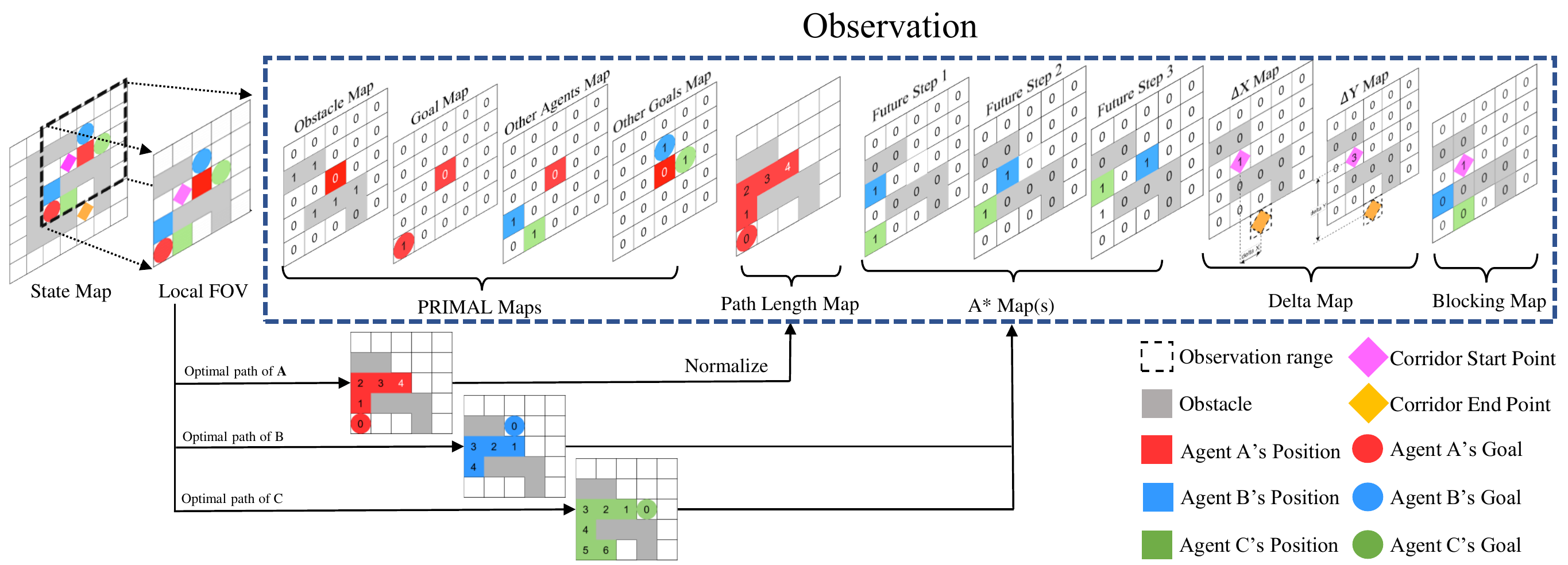}
  \vspace{-0.45cm}
  \caption{Observation space of the agents (here for agent A, in red), as detailed in Section~\ref{RAL2021-section:policy/observation}.
  The first four maps are identical to our previous work, providing information about obstacles, the agent's own goal, and nearby agents and their goals (e.g., agent B, in blue).
  The path length map displays the (normalized) shortest-path distance to the agent's own goal from all non-obstacle cells in the FOV.
  $n_{pred}$ (here, $3$) A* maps provide the future position of nearby agents, one per time step, predicted from single-agent A*.
  Finally, corridor information is encoded through the $\Delta X$, $\Delta Y$, and blocking maps.}
  \label{RAL2021-fig:whole_obs}
  \vspace{-0.3cm}
\end{figure*}


\subsection{Environment Setup}

In line with standard MAPF tasks, our environments are formulated as 2D discrete 4-connected grid worlds where agents, goals, and obstacles occupy one grid cell respectively. 
At each timestep, every agent can either move to a neighbouring location in one of the cardinal directions or wait at its current location (more details on the state/action spaces can be found in Section~\ref{RAL2021-section:policy}).
We consider highly structured worlds with moderate to high obstacle densities and long corridors which are created using a simple maze-generation algorithm parameterized by the world size, the average obstacle density, and the typical corridor length.
Corridors impart structure to the world, but they are also potential bottlenecks as they can lead to deadlocks, and prudent planning is required to efficiently navigate them.
A typical example of the worlds we consider can be found in Fig.~\ref{RAL2021-fig:map_demo}.
We consider two variants of MAPF: one-shot MAPF and lifelong MAPF (LMAPF).
While the focus of this work is on the lifelong variant of MAPF, we also considered a variant of one-shot MAPF to enable a discussion on solution quality and to ease comparison with baseline centralized~methods.

\subsection{One-shot MAPF}
\label{RAL2021-section:problem/oneshot_env}

In the one-shot MAPF variant, each agent is required to find a path to a unique goal assigned to it. 
Immediately upon reaching its goal, the agent disappears from the map and ceases to be a part of the state space of other agents.
Meanwhile, the unit grid cell occupied by the agent also frees up and can be accessed by other agents. An episode terminates when all agents have reached their goals.
Although uncommon, this MAPF formulation is valid when an agent can reach its goal and stay there without interfering with others, such as cars reaching a parking space~\cite{vsvancara2019online} or trains entering a station with parallel tracks~\cite{mohanty2020flatland,roost2020improving}.
In this variant, our goal is to minimize the makespan, i.e., the time needed for all agents to reach their goal.

\subsection{Lifelong MAPF}
\label{RAL2021-section:problem/continuous_env}

The LMAPF variant works in an online setting where agents do not have information about their subsequent tasks a priori, i.e., agents only know their current goal location and are assigned a new goal only upon arrival to their current one. 
These new goal locations are assigned randomly and are constrained to be some minimum Euclidean distance away from the agent's current goal.
The LMAPF environment can be run indefinitely or terminated after a certain desired number of timesteps.
The objective in LMAPF is the maximization of throughput, i.e., the average number of targets reached per unit of time.
Our constructed LMAPF environment aims to mimic real-world robot deployments in distribution centers, where robots are dynamically assigned new tasks and are constantly in motion to complete them.

\section{(L)MAPF as a RL Problem}
\label{RAL2021-section:policy}

In this section, we cast the (L)MAPF problem into the RL framework. In particular, we detail the observation and action spaces, the reward structure, and the policy network.


\subsection{Observation Space}
\label{RAL2021-section:policy/observation}

We consider a partially observable world where each agent can access the state of its surroundings within a limited square field-of-view (FOV) centered around itself (in practice, $11 \times 11$).
We believe that such a partially observable assumption is representative of real-world scenarios, where robots often only have access to incomplete information from their onboard sensors. 
Additionally, having a fixed, local FOV can help us learn a robust policy that can generalize to a wide range of world sizes while maintaining the same neural network structure. 

In this limited FOV, information is separated into several channels to aid learning.
Based on our previous works, four binary matrices provide information about obstacles, positions of other agents, goals of those observable agents, and the agent's own current goal position if within the FOV; three scalar values provide each agent with a unit vector pointing towards its goal and the absolute magnitude of the distance to its goal at all times~\cite{sartoretti2019primal}.
We also provide each agent with a path length map that contains the (normalized) shortest-path distance to its goal from each non-obstacle cell within its FOV.
These distances are calculated using single-agent A*, ignoring all other agents in the environment.
We believe that such a map resembles a gradient flow, enabling the agent to chart an effective (individual) trajectory even when its goal is not observable within the agent's FOV.

We further introduce three smaller spatial maps ($5 \times 5$ in practice, surrounded with zeros to reach $11 \times 11$) centered around the agents and encoding information about neighboring corridors.
Because of their narrow structure, corridors are potential bottlenecks in the world, and hence need to be efficiently navigated.
Corridors are regions where agents have at most two possible actions, excluding staying stationary.
Each corridor has two entry cells, barring corridors containing a dead-end that only have one.
We refer to these entry cells as \textit{Endpoints} and the cells outside the corridor connected to these endpoints as \textit{Decision Points}.
Decision points are named so, as agents occupying these cells have to take the critical decision of entering the corridor, which can potentially result in a future deadlock inside the corridor.
Fig.~\ref{RAL2021-fig:corridor} contains some examples of corridors and illustrates the special points discussed above.
All information about a specific corridor is encoded in the endpoint cells of that corridor.
The first two maps, namely the \textit{delta maps}, which contain values for $\Delta_X$ and $\Delta_Y$, provide a corridor's orientation as a displacement between the two endpoints of that corridor.
To construct these maps, the coordinates of the endpoints of a corridor are first obtained.
$\Delta_X$ is then defined as the difference of the x coordinates of these two endpoints and $\Delta_Y$ is the difference of the y coordinates of these two endpoints.
If a corridor is a dead-end, then it only has one endpoint and the delta values are 0.
The delta maps are very sparse and only contain non zero values in the grid cells which are the endpoints of a corridor.
For example, in Fig.~\ref{RAL2021-fig:corridor} (b), the delta maps will only contain non-values in the endpoint cells which are highlighted in blue. For the endpoint at the bottom-left, $\Delta_X = 2$ and $\Delta_Y = 2$.

The third map, namely the \textit{blocking map}, contains information about other agents currently within a corridor. 
Similar to the delta maps, the blocking map is a sparse map which only contains non-zero values at the endpoints of a corridor.
For an agent currently outside and close to a particular endpoint of a corridor, the blocking map contains a $1$ at that endpoint, if and only if there is at least one agent currently inside that corridor moving in a direction that would cause it to exit the corridor from that endpoint.
The conventions detailed in Section~\ref{RAL2021-section:learning/conventions} restrict agents from turning back in a corridor and thus, it is possible to ascertain the endpoint from which an agent would exit the corridor. 
For example, in Fig.~\ref{RAL2021-fig:whole_obs}, the red agent should not enter the corridor since the blue agent is moving towards the endpoint close to it and this would lead to a deadlock.
Consequently, the blocking map takes the value $1$ at the endpoint in the observation of the red agent. 

In addition to specific corridor data, we believe that agents can benefit from having an idea about other agents' future movements.
To this end, we let each agent construct a number, $n_{pred}$, of maps containing the predicted future position of other agents within its local FOV, one per map ($n_{pred}=3$ in practice).
In other words, $n_{pred}$ refers to the number of future timesteps that an agent looks ahead to.
For each timestep, the predicted future position of all visible neighbouring agents at that timestep is shown in the map.
These maps are generated using single-agent A*, under the assumption that each agent is alone in the world, and in practice would only require agents locally share goal information with their neighbors.
Thus, these maps are imprecise but can still provide considerable predictive power to the agent.

\subsection{Action Space}
\label{RAL2021-section:policy/actions}

We allow agents to take one out of five discrete actions in the grid world at every timestep: Moving one cell in any of the four cardinal directions or staying still.
During training, actions are sampled from a list of valid actions, and agents are prevented from taking invalid actions, examples of which are moving into another obstacle or agent.
Moreover, we also define some actions which fail to adhere to ideal predefined conventions about navigating corridors as invalid (detailed in Section~\ref{RAL2021-section:learning/conventions}).
A supervised loss function (i.e., valid loss) aids in learning valid actions.
We experimentally observed that learning valid actions does not depend on the preceding trajectory, and hence, bootstrapping with rewards can cause delayed and unfavorable convergence. Therefore, a supervised loss function works better in practice than rewards for learning the set of valid actions.
Additionally, to prevent convergence to oscillating policies that prevent exploration and stall learning, agents are not allowed to return to the location they occupied at the last timestep.
However, agents are allowed to stay still during multiple successive timesteps. 

\subsection{Reward Structure}

To motivate agents to reach their goals quickly, we penalize them at every timestep they are not on goal ($r_t = -0.3$), as is common in most reward functions for grid worlds.
Agents are also given a sizeable positive reward upon reaching their goal ($r_t = +5$), which effectively reinforces the immediate trajectories leading up to their goal.
Finally, although agents are not allowed to take invalid actions as discussed in Section~\ref{RAL2021-section:policy/actions}, it is still possible for them to collide with other agents in specific scenarios, such as two agents trying to move into the same empty cell.
In such cases, agents are given a collision penalty ($r_t = -2$).

\subsection{Network Structure}

Our work relies on the asynchronous advantage actor-critic (A3C) algorithm~\cite{mnih2016asynchronous}, and use the same network structure as our previous work~\cite{sartoretti2019primal},
parameterized by the set of weights~$\theta$.

The local observation channels are passed through two VGG-blocks~\cite{simonyan2014very}, followed by one last convolutional layer to finally obtain a one-dimensional vector of features.
In parallel, the goal unit vector and magnitude are first passed through one fully-connected (fc) layer.
The concatenation of both of these pre-processed inputs is then passed through two fully connected layers, and finally fed into a long-short-term memory (LSTM) cell. A residual shortcut~\cite{resnet} connects the output of the concatenation layer to the input layer of the LSTM. The output layers consist of the policy vector (discrete probability distribution over the $5$ possible actions considered) with softmax activation and the value.

The value output $V$ is updated to match the total long-term cumulative discounted return $R_r=\sum_{i=0}^{k}{\gamma^i r_{t+i}}$ at every visited state during the most recent episode, using a standard L2 loss $L_{value}$.
The policy gradient loss (training the actor output $\pi$ of the network) reads

\vspace{-0.4cm}
\begin{small}
\begin{equation}
    L_{actor} = \frac{1}{T} \sum_{t=1}^{T}{\sigma_H\cdot H(\pi(o_t)) - log \Big ( \pi \left (a_t|\pi,o;\theta \right ) A \left (o_t,a_t;\theta) \right ) \Big )  },
\end{equation}
\end{small}
\vspace{-0.5cm}

\noindent where $\sigma_H \cdot H(\pi(o_t)) = - \sigma_H \, \pi_t(a_t) \cdot \sum^{5}_{i=1}{ log( \pi_t(a_i) )}$, ( $\sigma_H = 0.01$ in practice), is an entropy term to encourage exploration and discourage premature convergence~\cite{babaeizadeh2016reinforcement}, and $A(o_t,a_t;\theta)$ an estimate of the advantage function (see Eq.\eqref{RAL2021-eq:advantage}).

As is standard in the advantage actor-critic algorithm, we use an approximation of the advantage function by bootstrapping using the value function (i.e., the output of the critic network):

\vspace{-0.4cm}
\begin{equation}
    A(o_t,a_t;\theta) = r_t + \gamma \, V(o_{t+1};\theta) - V(o_t;\theta)).
    \label{RAL2021-eq:advantage}
\end{equation}
\vspace{-0.6cm}

Besides the policy loss, we also rely on an additional loss to speed up the actor training, namely $L_{valid}$, which aims at reducing the log likelihood of selecting an invalid move. 

\vspace{-0.5cm}
\begin{equation}
    L_{valid} = \frac{1}{T} \sum_{t=1}^T { \sum_{i=1}^{5}{log(v_i(t)) \cdot \Tilde{\pi}_t(a_i) + log(1 - v_i(t)) \cdot (1 - \Tilde{\pi}(a_i))} },
    \label{RAL2021-eq:validLoss}
\end{equation}
\vspace{-0.4cm}

\noindent where $v_i(t)$ denotes the ground truth of action $i$'s validity at time $t$ ($1$ if valid, $0$ otherwise), and $\Tilde{\pi}$ is the result of a Sigmoid function being applied on $\pi$.

The final, combined training loss for the actor and critic outputs of the network reads $L_{final} = \alpha \cdot L_{value} + \beta \cdot L_{actor} + \zeta \cdot L_{valid}$, with $\alpha,\beta,\zeta \in \mathbb{R}$ manually tuned weights.


\section{Learning}
\label{RAL2021-section:learning}

In this section, we detail the methods used to achieve implicit agent coordination, and the actual training process.


\subsection{Coordination Learning}

In highly dense and constrained worlds with high traffic like those we consider, agents often find themselves in situations where coordination becomes necessary to find effective paths.
While centralized planners can achieve this explicitly by planning in the high dimensional joint space, decentralized policies require agents to learn coordination implicitly with limited information about the environment and without direct control over other agents' actions.
In the absence of any decentralized coordination learning, such as the techniques detailed in this section, we observed that agents distributively learn to act selfishly, merely trying to take the shortest A* paths to their goal and showing no coordinating behavior or regard for other agents' actions (even though coordination could lead to more optimal paths for themselves and a higher total reward for everyone).

In order to implicitly teach agents coordination, we use three techniques: \textbf{1)} identifying and forcing agents to learn certain conventions and exemplary behavior by using a supervised loss function (\textit{Convention Learning}), \textbf{2)} using expert demonstrations from centralized planners during training (\textit{Imitation Learning}), and \textbf{3)} sampling from a wide range of environments during training to enable learning of a robust, generalizable policy (\textit{Episode Randomization}).

\subsubsection{Convention Learning}
\label{RAL2021-section:learning/conventions}

In highly constrained worlds with a large number of long corridors, agents can drastically improve the quality of their paths if they learn a common policy that adheres to a certain set of conventions and effectively breaks symmetries~\cite{li2020new}.
We have identified certain conventions that are highly applicable to completely decentralized agents.
For example, an agent A should never enter a narrow corridor if another agent is currently moving inside that corridor in a direction opposite to that of agent A, as this will lead to a deadlock.
Similarly, agents moving inside a corridor should never reverse their movements abruptly and retrace their paths unless there is a deadlock (i.e., most other scenarios where agents follow such behavior are bound to be non-optimal).
If agents learn to follow the conventions above, they can navigate corridors much more efficiently and find higher quality paths with more coordinated movements.
However, these conventions are not evident to agents and are very difficult to learn using pure policy gradient methods, mainly because agents learn selfish policies, and rewards cannot effectively capture and reinforce such conventions.

To enable learning such conventions, we rely on our supervised valid loss function Eq.\eqref{RAL2021-eq:validLoss}, which teaches agents to avoid taking actions that go against the above conventions.
A metric called \textit{valid rate} keeps track of the fraction of actions chosen by agents which are valid, i.e., the success rate in selecting a valid action.
While the valid rate starts out low during training, agents are eventually able to learn to adhere to conventions and achieve a near perfect valid rate ($> 99.5\%$).
Interestingly, we observe that agents can also learn when they are inside corridors even though this information is not provided to them explicitly through the observation (i.e., only the endpoints of a corridor are evident in the observation).
We believe that this is made possible by the LSTM cell in the network architecture, and future work will explore the integration of more powerful recurrent networks with our current architecture.

\subsubsection{Imitation Learning}

The combination of RL and Imitation Learning (IL) has been shown to lead to faster convergence and higher quality solutions in robotic applications~\cite{zhu2018reinforcement,hester2017deep}.
In LMAPF, IL from centralized near-optimal planners which plan in the joint space can instill good quality coordination behavior in agents, which is challenging to accomplish by decentralized RL.
The ratio of RL to IL episodes is maintained close to $50\%$, as in our previous work.

Expert demonstrations in IL episodes are generated by the bounded suboptimal centralized planner ODrM* (with inflation $\epsilon=2$)~\cite{wagner2015subdimensional}, and a trajectory of observations and actions is attained.
We use these trajectories and the corresponding observations to minimize a standard behavior cloning loss:

\vspace{-0.35cm}
\begin{equation}
    L_{bc}= -\frac{1}{T} \sum_{t=0}^{T}{log \left( P (a_t|\pi,o_t;\theta) \right)}.
\end{equation}
\vspace{-0.35cm}

Since ODrM* is a one-shot MAPF planner, several one-shot MAPF instances need to be combined for a single LMAPF environment as is common when adapting one-shot planners to LMAPF~\cite{vcap2015complete,liu2019task}.
As a result, during training in the LMAPF environment, ODrM* is called at all timesteps where path replanning is required, i.e., all timesteps where at least one agent reaches its goal location.

\subsubsection{Environment Randomization}

To ensure that agents encounter diverse environments during training, we randomize the world size, density, and typical corridor length at the beginning of each episode.
Specifically, the size of our square worlds is uniformly sampled between $10$ and $70$, the average obstacle density between $20\%$ and $70\%$, and the typical corridor length between $3$ and $21$.
We find that uniformly sampling these parameters works well in practice.
Curriculum learning with increasing difficulty of environments has also been shown to be effective in practice~\cite{bengio2009curriculum}.
However, our experiments with implementations of curriculum learning did not yield significant performance improvements. Future works might investigate this technique.
In our environment randomization process, we believe larger-sized worlds are necessary for agents to learn to navigate to their goal even if it is a significant distance away, while the smaller sized worlds expose agents to cluttered scenarios which require coordination and collision avoidance to be solved effectively.
The positions of agents, obstacles, and goals are set randomly across the world, with the constraints that each agent has at least one path to its goal, and the goal is some minimum Euclidean distance away from the agent ($2$ cells in practice).
In addition to this, the agents' initial positions are constrained, such that there is no more than one agent inside any narrow corridor.
This is primarily done to ensures that the conventions discussed in Section~\ref{RAL2021-section:learning/conventions} are adhered to since the beginning of the episode.


\subsection{Training}
\label{RAL2021-sec:training}

\subsubsection{General Training Parameters}

In line with standard RL parameter choices, we use a discount factor ($\gamma$) of $0.95$ and an episode length of $256$.
However, IL episodes have a length of $64$ because of the high cost associated with repeatedly calling ODrM*.
In addition to performing a gradient update at the end of the episode, we also perform one immediately after an agent reaches its goal.
As a result, an agent may be trained more than once per episode, depending on the number of targets it reaches.
We use the NAdam Optimizer with learning rate \(2.10^{-5}\) and decay the learning rate proportionally to the inverse square root of the episode count.

\subsubsection{Distributed Training Framework}

We train our model utilizing Ray, a distributed framework for machine learning~\cite{moritz2018ray}.
Ray allows us to bypass Python's Global Interpreter Lock and easily scale to a cluster using multiple GPUs.
In practice, the final policy was trained on a single workstation equipped with a i9-10980XE CPU ($18$ cores, $36$ threads) and one NVIDIA Titan RTX GPU.
The code employs $9$ remote training nodes, $4$ of them calculating gradients via imitation learning with ODrM*, while the other $5$ run pure RL episodes using the most up-to-date policy.
The choice of these numbers has been made experimentally to keep the RL to IL episodes ratio close to $50\%$.

Each node is equivalent to a single meta-agent of the overall A3C architecture and contains a copy of the LMAPF environment in which $8$ agents are learning to plan paths.
All $9$ nodes run in parallel and pass gradients to the master node to be applied to the global network asynchronously.
Training lasts around $10$ hours and converges within $35k$ episodes (nearly $10$x fewer episodes and $24$x shorter training time than our previous work).
The full training code for PRIMAL$_2$ is available at \url{https://bit.ly/PRIMAL2}, and can easily be adapted to other MARL tasks.


\section{Results}
\label{RAL2021-section:results}

This section presents our one-shot and LMAPF results comparing PRIMAL$_2$ to state-of-the-art planners.


\subsection{One-Shot MAPF Results}

For all of our experiments (one-shot and LMAPF), we systematically tested team sizes in $\{4,8,16,....1024\}$, world sizes in $\{20,40,80,160\}$, densities in $\{0.3, 0.65\}$, and typical corridor lengths in $\{1,10,20\}$.
We run $50$ tests for each possible combination of the above parameters, barring a few infeasible combinations, and average the results in our plots.
Specifically, we do not run tests containing $64$ agents or more in $20$-sized worlds, $256$ agents or more in $40$-sized worlds, and $1024$ agents in $80$-sized worlds.
To eliminate any bias in our results, all planners encounter the same test scenarios.
We use makespan (i.e., time until all robots are on target) and success rate as our primary evaluation metrics.

We select CBSH-RCT~\cite{li2020new} as our optimal planner and ODrM* (with $\epsilon = 10$)  as our bounded suboptimal centralized planner~\cite{wagner2015subdimensional}, with a timeout of $60$s to remain consistent with other works in the field.
We also use PRIMAL as a baseline MARL-based decentralized planner~\cite{sartoretti2019primal}.
For PRIMAL and PRIMAL$_2$, we allow a maximum of $320$ timesteps for $20$ and $40$-sized worlds, $480$ timesteps for $80$-sized worlds, and $640$ timesteps for $160$-sized worlds.
We trained separate dedicated models for one-shot and LMAPF for these planners.
Note that, while centralized planners have access to the full state of the system, agents in PRIMAL and PRIMAL$_2$ only have access to a limited FOV.

One of our recent studies into PRIMAL showed that unsuccessful episodes still often drive an overwhelming majority of agents to their goal~\cite{sartoretticombined}.
Therefore, in our one-shot MAPF testings, we further provide PRIMAL/PRIMAL$_2$ results where we consider an episode to be successful when $100\%$ and $95\%$ of agents reach their goals successfully.
The primary motive of adding the $95\%$ success metric is to better gauge the performance of decentralized planners like PRIMAL$_2$, and is not meant to replace the standard $100\%$ binary metric that remains the norm in one-shot MAPF. 
In order to ensure fairness, we also adapted centralized planners to the $95\%$ success metric.
At the start of each episode, we sample a subset of agents which consists of only $95$\% of the total agents.
We remove the remaining agents from the map and plan for the reduced subset of agents only.
We run 10 such iterations for every episode (i.e, with 10 different subsets of $95$\% of the agents) and classify that episode as a success if the planner is able to successfully find a solution in any of the 10 runs.
All result plots are available at \url{https://bit.ly/PRIMAL2} and in the supplemental material, including LMAPF results.
Fig.~\ref{RAL2021-fig:results-oneshot} shows the success rates and path lengths in a representative~case.

Based on our results, we first notice that all planners have very high success rates in worlds with low densities and short corridors.
Next and as expected, we observe that the performance of all planners drops with increasing obstacle density and corridor lengths.
This is because highly constrained environments have increased conflict in agent paths and require extensive inter-agent coordination to solve effectively.
While all planners perform nearly equivalently in small team sizes (up to $16$ agents), we observe that PRIMAL$_2$ with success metric $100\%$ is slightly outperformed by centralized planners in moderate team sizes ($16$-$128$ agents).
In large team sizes ($> 128$ agents), the performance of centralized planners drops sharply, which is a common problem faced by many centralized planners, due to the exponential increase in the dimension of the joint configuration space to be searched.
As a result, PRIMAL$_2$ is able to outperform centralized planners as team sizes scale above $128$ agents. 
We also note that PRIMAL$_2$ also comfortably outperforms our previous work, PRIMAL, in both moderate and large team sizes.
Interestingly, we observe that while the performance of PRIMAL$_2$ with $100\%$ success metric drops off gradually as the number of agents are increased, that of PRIMAL$_2$ with $95\%$ success metric stays nearly constant.
Upon careful inspection of the results, we often find that a few agents get stuck in undesirable looping behaviour which could be corrected with the introduction of a centralized planner at this stage, and will be the focus of future works.
Thus, although PRIMAL$_2$ is outperformed by centralized planners in moderate team sizes when using the standard $100\%$ metric, we believe that the correction of this looping behaviour can significantly increase success rates and it will be the focus of future works.

\begin{figure}[t]
\vspace{0.2cm}
\begin{center}
\includegraphics[width=0.96\linewidth]{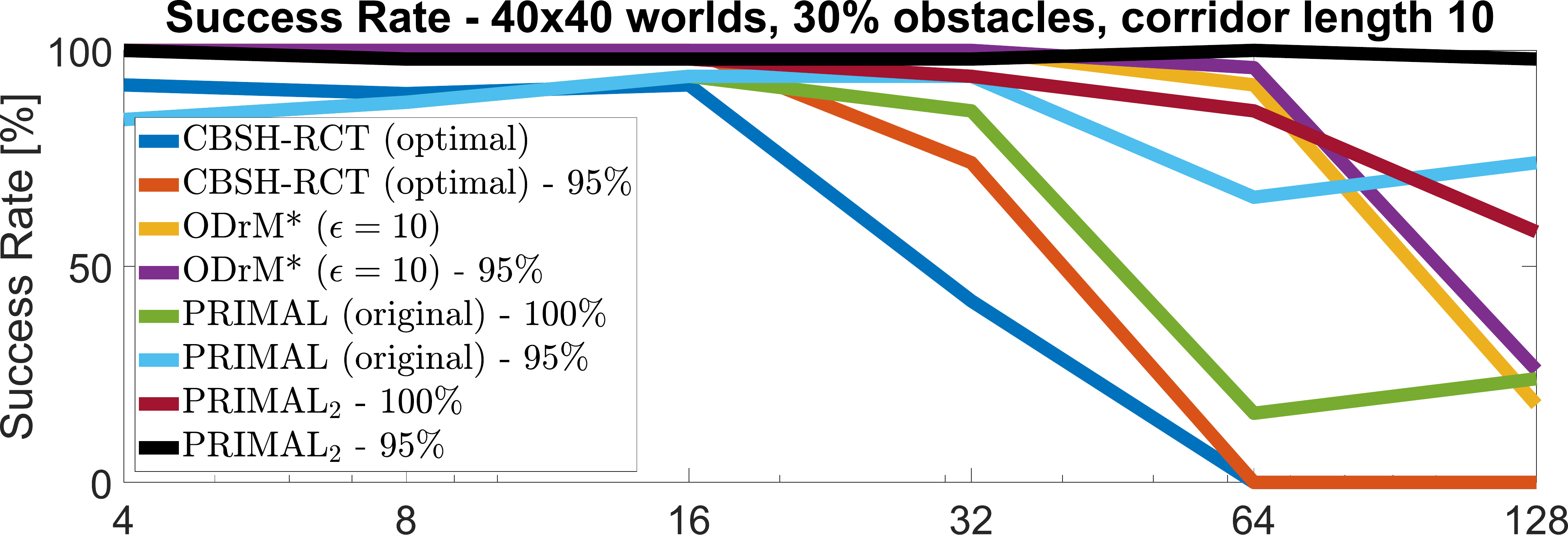}  \\[0.1cm]
\includegraphics[width=0.96\linewidth]{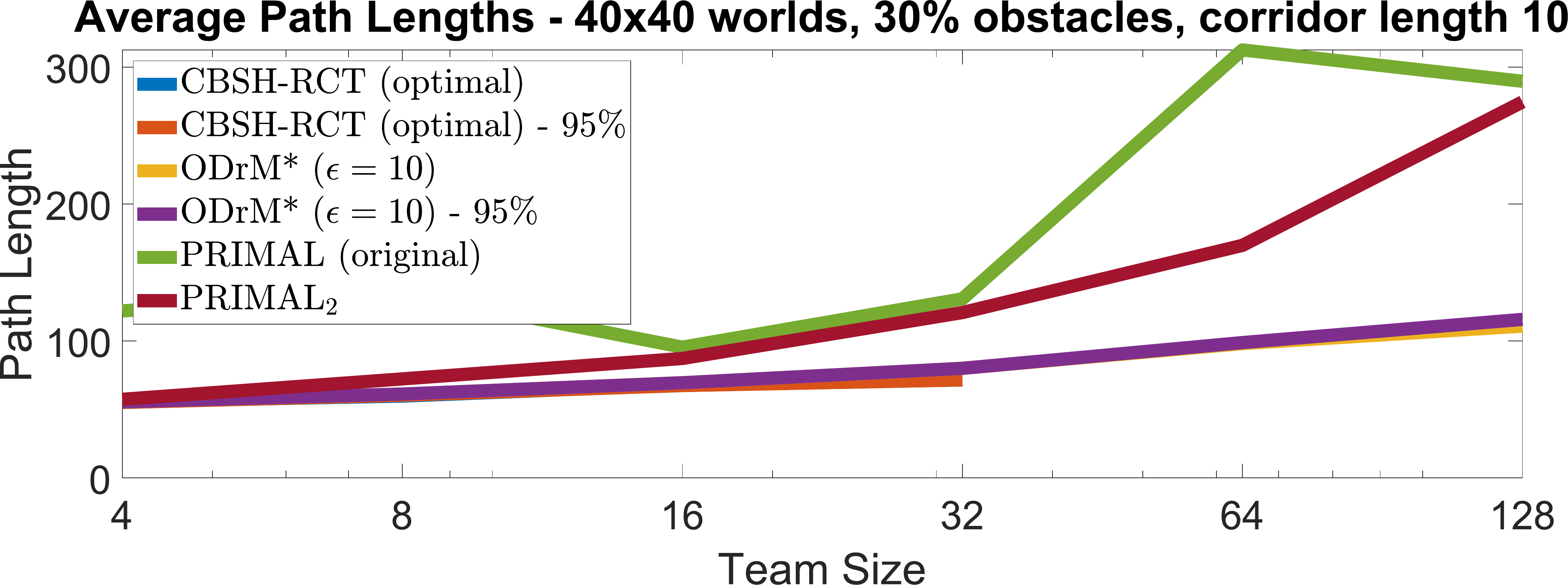}
\end{center}
\vspace{-0.6cm}
\caption{Success rate and plan lengths of the considered planners in a representative one-shot MAPF scenario.
As expected, PRIMAL$_2$ is slightly outperformed by centralized planners in smaller teams, but outperforms them in larger teams ($\geq 256$ agents). Due to its decentralized nature, PRIMAL$_2$ trajectories are considerably longer than centralized planners.}
\label{RAL2021-fig:results-oneshot}
\vspace{-0.5cm}
\end{figure}

Due to its decentralized and incomplete nature, we find that paths yielded by PRIMAL$_2$ tend to be considerably longer than ODrM* and CBSH-RCT on average.
In general, solution quality tends to decrease as the number of agents increase and as the environments become more constrained with longer corridors and higher obstacle densities.
We observe that, in team sizes of up to 32 agents, the differences in solution quality are minute, where PRIMAL$_2$ paths tend to be $25\%$- $50\%$ longer than ODrM* on average.
In team sizes between 64 to 256 agents, the paths yielded by PRIMAL$_2$ further decrease in quality due to the increase in scale and on average, are $75\%$-$125\%$ longer than ODrM*.
ODrM* is unable to generate solutions for $512$ agents and above, and thus it is not possible to get an estimate of the path suboptimality for the largest teams.
Intuitively, we expect the solution quality of PRIMAL$_2$ in comparison to the optimal solution to drop down even further in large teams.
It is interesting to note that in comparison with our previous work, PRIMAL, the paths of PRIMAL$_2$ tend to be considerably shorter which we believe is a result of the extensive inter-agent coordination achieved in PRIMAL$_2$. 
Concluding, even though we observe a significant increase in implicit agent coordination compared to our previous work, the decentralized nature of PRIMAL$_2$ makes it very hard to achieve perfect joint coordination with the same quality of paths as other centralized planners.


\subsection{LMAPF Results}

\begin{figure}[t]
\vspace{0.2cm}
\begin{center}
\includegraphics[width=0.96\linewidth]{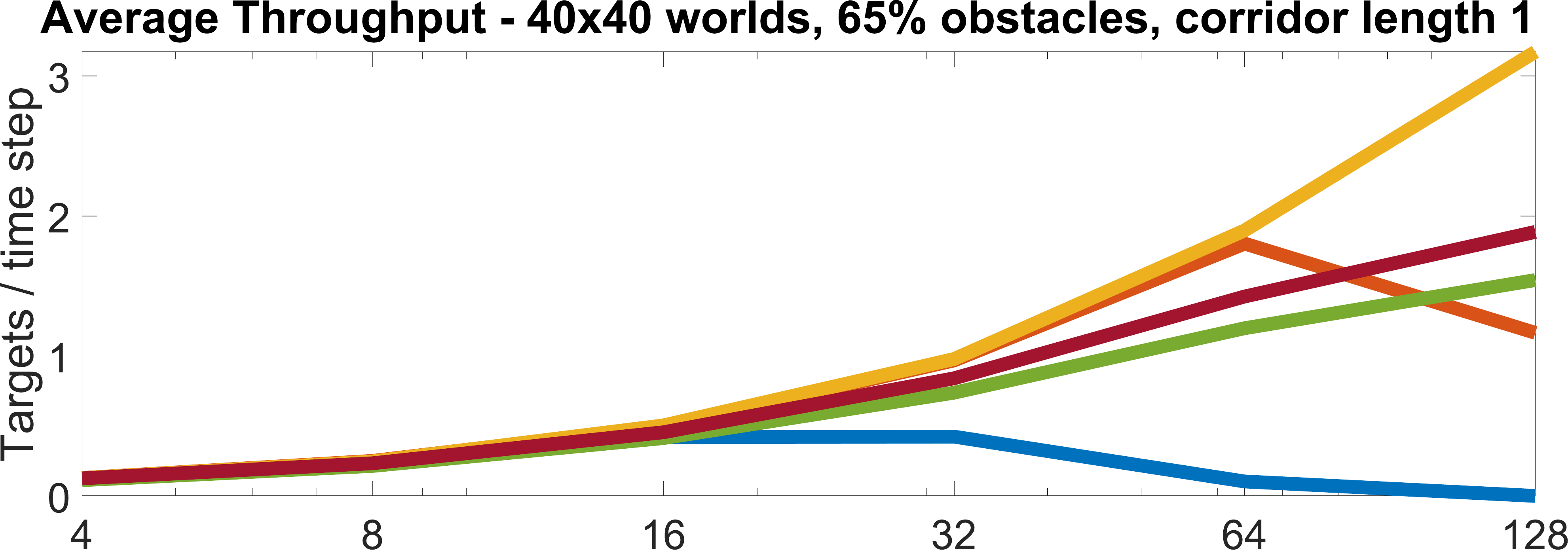} \\[0.2cm]
\includegraphics[width=0.96\linewidth]{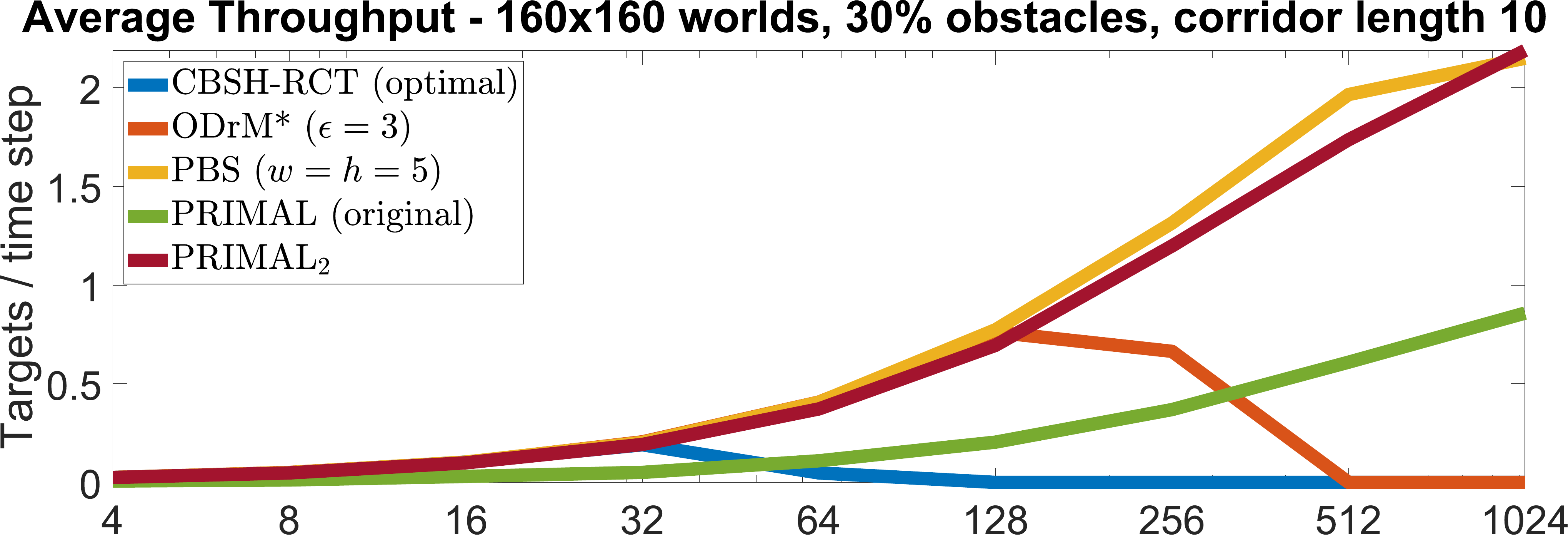}  \\[0.2cm]
\includegraphics[width=0.96\linewidth]{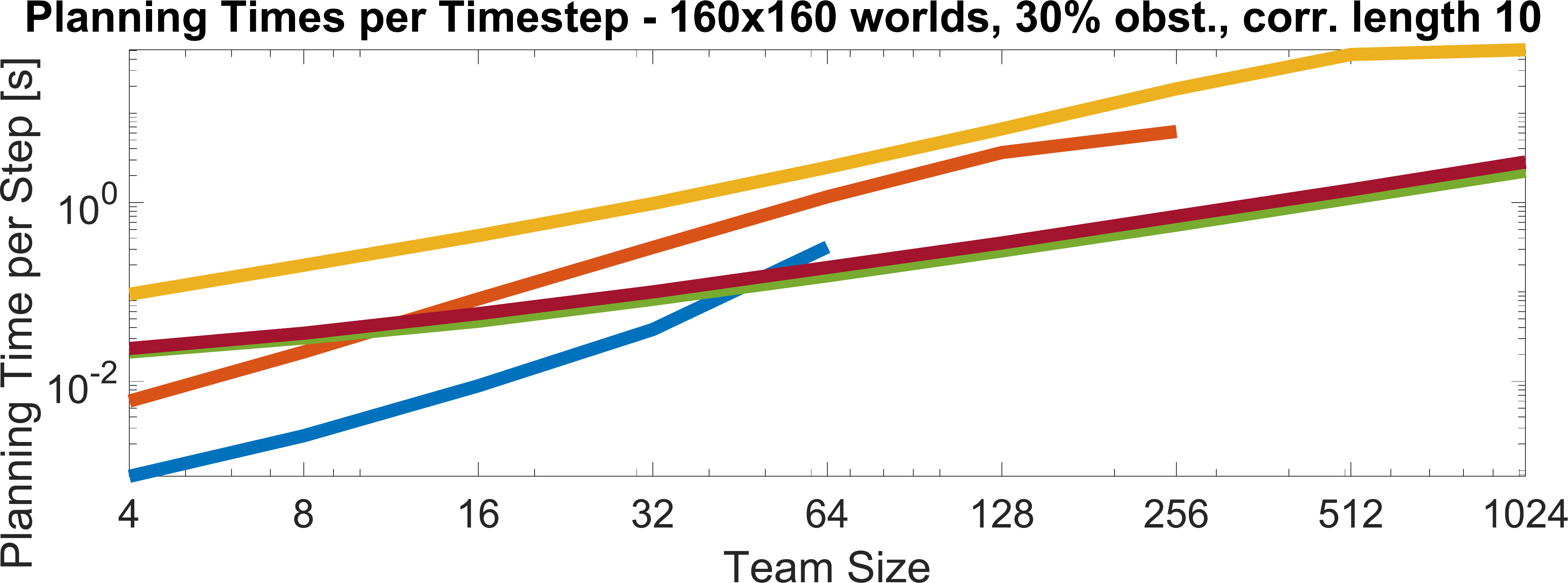}
\end{center}
\vspace{-0.55cm}
\caption{Average throughput and plan lengths of planners for LMAPF; interrupted lines indicate that no solutions were found for larger team sizes.
Although windowed-PBS is able to match or outperform PRIMAL$_2$, it plans an order of magnitude slower, with individual replanning instances reaching a minute for larger teams.
Note how the new coordination techniques introduced in this work allow PRIMAL$_2$ to significantly surpass PRIMAL.
}
\label{RAL2021-fig:results}
\vspace{-0.3cm}
\end{figure}

In LMAPF, agents aim at continually planning paths online and maximizing the throughput (i.e., the average number of targets reached per timestep).
To implement conventional baselines for LMAPF, we decompose the problem into a series of one-shot MAPF instances as is commonly done ~\cite{wan2018lifelong}.
We select CBSH-RCT as a constrained-environment-optimized optimal planner~\cite{li2020new}, as well as ODrM* (with $\epsilon=3$) and Windowed-PBS (with w=$5$ and h=$5$) as our bounded suboptimal planners~\cite{wagner2015subdimensional,li2020lifelong} and use a timeout of $60$s per (re-)planning instance.
For all planners, our experiments last 128 timesteps in $20$- and $40$-sized worlds, $192$ timesteps in $80$-sized worlds, and $256$ timesteps in $160$-sized worlds.
For the conventional baselines, we compute the team's average throughput until that maximum number of timesteps, or until one (re-)planning instance times out, whichever happens first.
By doing that, we note that early timeouts do not impact the throughput negatively.
Fig.~\ref{RAL2021-fig:results} presents our results in two representative scenarios.

We first observe that both centralized and decentralized planners have high throughput in worlds with low density, small team sizes, and short corridors.
PRIMAL, PRIMAL$_2$, and windowed PBS scale remarkably well to larger teams, while the performance of ODrM* and CBSH-RCT drops sharply above $128$ agents.
However, as the typical corridor length and the average obstacle density is increased to make the worlds more constrained and challenging, we see drops in performance for all planners.
In general, we observe that windowed PBS is able to marginally outperform PRIMAL$_2$ for nearly all scenarios up to $512$ agents.
While windowed-PBS can still generally handle $1024$-agent scenarios, we start to see timeouts in some episodes.
More generally, we observe that the re-planning times of windowed-PBS are well over an order of magnitude higher than PRIMAL$_2$ for moderate and large team sizes.
It is worth noting that to adapt windowed-PBS to our problem definition, we need to re-plan at every timestep an agent reaches its goal, which can happen nearly every time step for larger teams.
In this context, we note that PRIMAL$_2$ offers real-time re-planning capabilities (with generally sub-second decentralized re-planning), which might make it more attractive for online deployments.

More generally, our results highlight the trade-off between maximizing throughput and minimizing planning time. 
While windowed PBS is able to achieve high throughput for team sizes up to $512$ agents through frequent re-planning and larger planning times, PRIMAL$_2$ achieves slightly more moderate throughput but plans drastically faster.
We also note that, while PRIMAL and PRIMAL$_2$ have equivalent performance in worlds with low densities and short corridors, PRIMAL$_2$ easily outperforms PRIMAL in larger team sizes ($\geq 128$ agents) and in constrained worlds with long corridors.
We believe the additional coordination learning techniques introduced in this paper are the reason for this improvement. Interesting learned maneuvers can be observed in the videos of our PRIMAL$_2$ results, which can be found at
\url{http://bit.ly/PRIMAL2videos}.

\begin{table*}
\caption{LMAPF throughput on moving.ai benchmarks (timeout: 60s, result marked as "-" if all scenarios time out)}
\begin{center}
\begin{tabular}{ |c|c|c|c|c|c|c| } 
\hline
Planner & Maze (4-64) & Maze (128-512) & Maze (1024-2048) & Warehouse (4-64) & Warehouse (128-512) & Warehouse (1024-2048) \\
\hline
ODrM* ($\epsilon=3$) & 0.21 & - & - & 0.17 & - & - \\
PRIMAL & 0.07 & 0.31 & 0.71 & 0.08 & 0.74 & 2.10 \\
Windowed-PBS & \textbf{0.26} & \textbf{1.08} & - & \textbf{0.18} & \textbf{1.95} & - \\
PRIMAL$_2$ & 0.17 & 0.55 & \textbf{1.05} & 0.17 & 1.49 & \textbf{3.36} \\
\hline
\end{tabular}
\end{center}
\label{RAL2020-table:scaling}
\vspace{-0.6cm}
\end{table*}

We further tested the above LMAPF baselines on the warehouse and maze maps available on movingai.com~\cite{stern2019mapf}, which serve as benchmark maps for the MAPF community.
These results are summarized in Table ~\ref{RAL2020-table:scaling}. 
For ease of analysis and conciseness, we present averaged results for small ($4$-$64$ agents), moderate ($128$-$512$ agents) and large teams ($1024$-$2048$ agents).
In the maze maps, we find that windowed-PBS is able to outperform PRIMAL$_2$ in small and moderate team sizes, but falls behind in large team sizes where it generally times out.
These results follow a similar pattern to the results on our LMAPF environments, wherein centralized planners are able to outperform or perform on par with PRIMAL$_2$ in small and moderate team sizes, but fall behind PRIMAL$_2$ in large teams.
In the warehouse maps, we observe that PRIMAL$_2$ performs on par with windowed-PBS and ODrM* in small teams.
However, windowed-PBS is able to perform better in moderate team sizes for these maps.
Similar to the maze maps, windowed-PBS times out in large team sizes while PRIMAL$_2$ continues to increase throughput.
We also note that both for maze and warehouse maps, PRIMAL$_2$ performs adequately in $2048$ agent scenarios, the largest team size we have tested yet.

\begin{figure}[t]
\vspace{0.2cm}
\begin{center}
\includegraphics[width=0.96\linewidth]{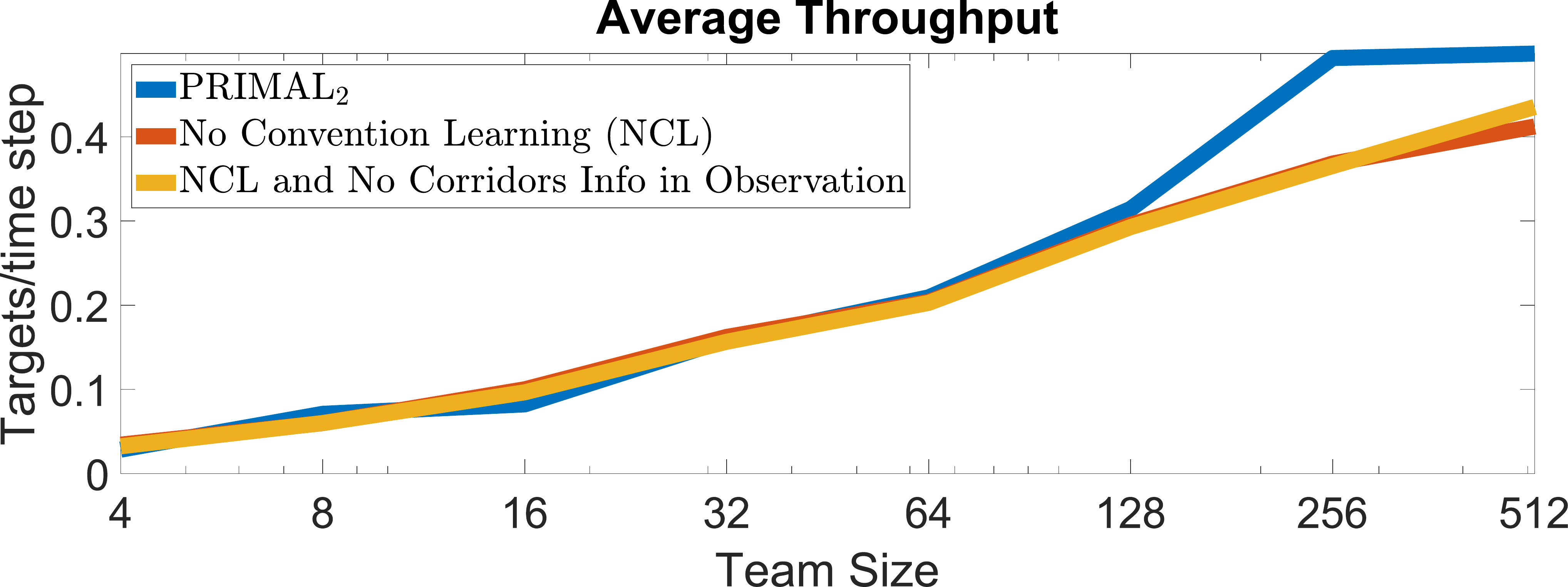}
\end{center}
\vspace{-0.5cm}
\caption{Performance of PRIMAL$_2$, compared to two PRIMAL$_2$ variants: 1) no convention learning, and 2) no convention learning, and no corridor information in the observation.
Note how, in larger teams ($\geq 128$ agents), PRIMAL$_2$ outperforms these variants by more than $10\%$ in terms of throughput, showcasing the importance of convention learning.}
\label{RAL2021-fig:continuous_PRIMAL$_2$_baselines}
\vspace{-0.3cm}
\end{figure}

Finally, we also present averaged results over all worlds, which compare the performance of PRIMAL$_2$ to two other PRIMAL$_2$ variants: 1) no convention learning, and 2) no convention learning and a reduced observation with no corridor information channels in the agents' observation (Fig.~\ref{RAL2021-fig:continuous_PRIMAL$_2$_baselines}).
We observe that, while these three planners perform near-identically up to $64$ agents, PRIMAL$_2$ surpasses them in larger team sizes (by around $10\%$).
We believe that these results show that for smaller teams, the gains from following conventions are neutralized by the higher freedom of movement enjoyed by not following any.
However, as team sizes increase, these conventions become integral to plan effectively and bring order to agents' movements.

\section{Conclusion} \label{RAL2021-section:conclusion}

This work introduced PRIMAL$_2$, a new distributed reinforcement learning framework for lifelong multi-agent path finding in highly constrained worlds.
In this framework, agents plan individual paths online in a wholly decentralized way, based on local information and interaction toward exhibiting joint maneuvers.
We focused on achieving implicit agent coordination by helping agents learn ideal behaviour through conventions, which effectively break symmetries and bring harmony to their movement.
Through our results, we highlighted the importance of these conventions in larger teams and experimentally showed that PRIMAL$_2$ agents are successful at learning them.
We also showed that PRIMAL$_2$ scales to arbitrarily large teams, up to $2048$ agents, and can plan effective paths online while producing throughput comparable to centralized planners for both one-shot and LMAPF.
Future work will try to further improve implicit agent coordination via a variety of techniques such as more powerful recurrent network architectures, the systematic investigation of RL-to-IL ratios, or the use of recent off-policy learning methods.


\section*{ACKNOWLEDGMENTS}
\label{RAL2021-acknowledgments}

\noindent 
We would like to extend our warmest gratitude to Jiaoyang Li, for happily agreeing to share her code
and providing instrumental support for the CBSH-RCT and PBS algorithms used as baselines.
Detailed comments from anonymous referees contributed to the presentation and quality of this paper.


\bibliographystyle{IEEEtran}
\bibliography{refs_full}

\end{document}